\documentclass[10pt, a4paper]{article}
\usepackage{lrec2022} % this is the new LREC2022 Style
\usepackage{multibib}
\newcites{languageresource}{Language Resources}
\usepackage{graphicx}
\usepackage{tabularx}
\usepackage{soul}

\usepackage{titlesec}
\titleformat{\section}{\normalfont\large\bf\center}{\thesection.}{1em}{}
\titleformat{\subsection}{\normalfont\SmallTitleFont\bf\raggedright}{\thesubsection.}{1em}{}
\titleformat{\subsubsection}{\normalfont\normalsize\bf\raggedright}{\thesubsubsection.}{1em}{}
\renewcommand\thesection{\arabic{section}}
\renewcommand\thesubsection{\thesection.\arabic{subsection}}
\renewcommand\thesubsubsection{\thesubsection.\arabic{subsubsection}}
\usepackage{epstopdf}
\usepackage[utf8]{inputenc}

\usepackage{hyperref}
\usepackage{xstring}

\usepackage{color}

\title{Connecting a French Dictionary from the Beginning of the 20th Century to Wikidata}

\name{Pierre Nugues}

\address{Lund University \\
         Lund, Sweden \\
         pierre.nugues@cs.lth.se\\
         \textit{Paper originally published in the Proceedings of the Language Resources and Evaluation Conference, 2022}}

\abstract{
The \textit{Petit Larousse illustré} is a French dictionary first published in 1905. Its division in two main parts on language and on history and geography corresponds to a major milestone in French lexicography as well as a repository of general knowledge from this period. Although the value of many entries from 1905 remains intact, some descriptions now have a dimension that is more historical than contemporary. They are nonetheless significant to analyze and understand cultural representations from this time. A comparison with more recent information or a verification of these entries would require a tedious manual work. In this paper, we describe a new lexical resource, where we connected all the dictionary entries of the history and geography part to current data sources. For this, we linked each of these entries to a wikidata identifier. Using the wikidata links, we can automate more easily the identification, comparison, and verification of historically-situated representations. We give a few examples on how to process wikidata identifiers and we carried out a small analysis of the entities described in the dictionary to outline possible applications. The resource, i.e. the annotation of 20,245 dictionary entries with wikidata links, is available from GitHub (\url{https://github.com/pnugues/petit_larousse_1905/}). \\ \newline  }
\begin{document}

\maketitleabstract

\section{Introduction}
The \textit{Petit Larousse illustré} is a one-volume dictionary of French, first published in 1905. Its ambition was as much cultural as linguistic with numerous illustrations and encyclopedic developments accompanying the definitions of words, things, and people. It was also a truly popular dictionary with, for instance, a sense order that followed the sense frequency and not the word history. As a consequence, this dictionary was a true commercial success and became ubiquitous in French homes. From its first edition, this dictionary has been annually updated until today and has kept a far-reaching cultural influence.

\subsection{The \textit{Petit Larousse Illustré}}
%\subsection{Structure}
The \textit{Petit Larousse illustré} stems from two main previous encyclopedias: A large one, the \textit{Grand dictionnaire universel du XIXe siècle} in 15 volumes and two supplements, published between 1866 an 1876 and a shortened and illustrated version of it: The \textit{Nouveau Larousse illustré} in seven volumes and a supplement from 1897 to 1904. 

While an offspring of them, the \textit{Petit Larousse illustré} has a completely new structure and format: It consists of one compact and portable volume, divided into three parts. These parts are of unequal sizes: 
\begin{enumerate}
    \item The first one, \textit{langue française}, is a   dictionary of French with some encyclopedic content. It is restricted to the common nouns, verbs, adjectives, adverbs, and grammatical words (1066 pages); 
    \item The second part, \textit{locutions}, contains quotes, mostly from classical Latin authors. It is much smaller than the two other parts (32 pages); and
    \item The third part, \textit{histoire et géographie}, contains short encyclopedic descriptions of people, countries, locations, intellectual or art works (660 pages).
\end{enumerate}

The third part, often called the ``proper nouns,'' has 20,245 entries. The dictionary was designed to be a best-seller that should find its place in all the French homes. The Larousse editors selected a content that they believed would correspond to a core cultural knowledge at the beginning of the 20th century: What the readers should know and what they wanted to know. As a consequence, it is also a snapshot the world's view and a repository of the cultural values of France at that time. 

\subsection{Entry Linking}
The printed content of such a dictionary is by nature static. Elementary information on the gender or country of birth of persons is sometimes missing; their occupations are difficult to extract or incomplete; the exact geographical coordinates of locations are missing from the dictionary; etc. We can retrieve such information from other sources, but connecting and relating the entries to current knowledge sources in a machine-readable format can be a time-consuming and tedious task. 

This work tries to bridge this gap by connecting all the entries from the history and geography part in the dictionary to wikipedia information (wikidata items). 

Beyond augmenting the dictionary with additional encyclopedic data, the wikidata links also enable us to identify and verify information, and compare historically-situated representations with currently available descriptions from Wikipedia and other sources.

%We also describe how we can use these items to enrich the dictionary with external knowledge.

Using the wikidata identifiers, we describe a few application examples. We show how we can process information for each entry automatically and complement the Larousse text with a wealth of external, structured, and recent knowledge. 

\subsection{Contributions}
As main contribution of this paper, 
\begin{enumerate}
    \item We describe how we linked all the entries of the third part of the dictionary, more than 20,000, to unique wikidata identifiers;
    \item We conducted a small, but significant study on the interannotator agreement on the links;
    \item We give a few examples of applications we can build with this resource; and finally
    \item We release the annotated dataset as a public language resource.
\end{enumerate}
 
The annotated dataset is available in the JSON format as a supplementary material. The file, \url{larousse_1905_wd.json}, can be downloaded from this address: \url{https://github.com/pnugues/petit_larousse_1905/}.

\section{Previous Work}
\subsection{Digital Editions of the \textit{Petit Larousse Illustré} from 1905}
To the best of our knowledge, there are only two completed public digital editions of historic versions of the \textit{Petit Larousse illustré:}
\begin{enumerate}
    \item \newcite{manuelian2009} transcribed the first part of the dictionary in a machine-readable format, where they segmented each entry into structured fields. These fields followed the Text Encoding Initiative guidelines for dictionaries \cite{TEI2021};
    \item \newcite{bohbot2018} started from the first project and expanded it. The authors completed the digital edition of the whole volume from 1905 as well as selected editions from 1905 to 1948. The entries are also structured according to the TEI. Users can consult and compare these editions on the Nénufar web site\footnote{\url{http://nenufar.huma-num.fr/}}.
\end{enumerate}

Digital editions of dictionaries, even when they enrich the text from static printed versions, do not connect the words to external knowledge. This is also the case with the current versions of the digital \textit{Petit Larousse illustré} from 1905. This lack of connectivity obviously impairs possible openings to contemporary knowledge. 

\subsection{Entity Annotation}
Most datasets with an entity annotation are text corpora, where the mentions of entities are bracketed and annotated with a unique identifier, i.e. all the mentions of named entities in the text are linked to an identifier. As identifiers, the most popular nomenclatures are Wikipedia pages, Freebase \cite{Bollacker2008}, or Wikidata.

AIDA CoNLL \cite{hoffart2011} is one such dataset. It builds on the CoNLL 2003 corpus annotated with named entity mentions \cite{conll2003}. The authors extended this corpus by linking all the mentions to YAGO2 names and, subsequently, to wikidata numbers. 

The TAC 2017 entity discovery and linking task dataset \cite{TAC2017} is a trilingual corpus of English, Spanish, and Chinese texts annotated with named entity mentions and links to Freebase. This dataset also contains a \textit{silver-standard} annotation of texts in ten low-resource languages.

While most entity annotation focuses on contemporary texts, \newcite{Hamdi2021multilingual} used a corpus of historical newspapers between 1850 and 1950 in four languages. The authors annotated 6700 mentions of named entities that they linked to wikidata.

\subsection{Entity Linking and Entity Descriptions}
\newcite{Logeswaran2019} showed the value of dictionaries of entities in the automatic task of zero-shot entity linking for specialized domains. An entry in their dictionary consists of the name of the entity, a definition of it, and a unique identifier. \newcite{DBLP:conf/aaai/NieCWLP18} and \newcite{chen-etal-2019-enteval} also obtained performance improvements in English, and \newcite{botha-etal-2020-entity} in multiple languages, when incorporating entity descriptions in the entity linking task. \newcite{ma-etal-2021-muver} extended this approach with multiple descriptions. 

In this paper, we describe a resource similar to the descriptions these authors used, but tied to dictionary definitions from the beginning of the 20th century.  

%where we connects all the definitions of the entries in the history and geography part of the \textit{Petit Larousse illustré} to unique wikidata items. 

\subsection{Annotation of Entries in Historical Encyclopedias}
Finally, to the best of our knowledge, there is no peer-reviewed paper, or even technical document, on a systematic linking of entries of historical encyclopedias  to wikidata. In its graph of properties, Wikidata includes links to biographies such as the \textit{Allgemeine Deutsche Biographie}, with more than 24,100 links, or encyclopedias, such as the \textit{Enciclopedia Treccani}, with about 23,300 links, but with no information that would describe these projects.

% décrit par P1343
% Deutsche Bio
% ?item wdt:P1343 wd:Q590208.
% 24166 out of 26 500
% identifiant Trecanni
% ?item wdt:P3365 ?id.
% 23296
% identificant Deutsche Biographie
% pas Allgemeine Deutsche Biographie
% P7902
% Liste des bios: ?item wdt:P31 wd:Q97584729 (130)
% Liste des encyclopédies: Q55452870 (262)

\section{Wikidata}
As connection points and unique identifiers, we used wikidata items. Wikidata is a data graph originally designed to handle the multilingual versions of wikipedia, such as French and Italian. For this, wikidata assigns a unique number to each thing that is the subject (title) of a wikipedia article in any language. Whatever the language, the identification number is the same.

For example, Pierre Larousse, the editor of the \textit{Grand dictionnaire universel du XIXe siècle}, has the wikidata number \verb=Q313709=, which has links to the French, English, Italian, Russian, or Chinese versions of wikipedia. In addition to links to wikipedia articles, wikidata contains structured information which will depend on the category (class) of the thing described: Pierre Larousse is an instance of human; it therefore has a date and place of birth, death, etc. Depending on their notoriety, some wikidata items contain a lot of information, others less or none. 

For Pierre Larousse, this information is available from the page: \url{https://www.wikidata.org/wiki/Q313709}, as well as in the JSON and Turtle formats, respectively \url{https://www.wikidata.org/wiki/Special:EntityData/Q313709.json} and \url{https://www.wikidata.org/wiki/Special:EntityData/Q313709.ttl}.

Wikidata also links wikipedia articles with other sites in the wikimedia galaxy, like wikisource or commons. Finally, many wikidata elements have links to external databases, such as authority lists from the Library of Congress, the Deutsche Nationalbibliothek, the National Library of France, or SUDOC, which allows more information to be extracted.

%However, this ideal situation suffers from a few exceptions, as some wikidata elements have no external link, have a bad structure or are wrong. 

\section{Method}
\label{sec:method}
We used the digitized version of the \textit{Petit Larousse illustré} available from the Nénufar website \cite{bohbot2018}. The history and geography part has 20,245 entries in total and we manually connected each of them to one or more wikidata items. Although the XML-TEI structure of the entries contains interesting information, we only considered the raw text.

Some entries are easy to link like the person names. The entry:
\begin{quote}
AALI-PACHA, homme d'Etat turc, né à Constantinople. Il a attaché son nom à la politique de réformes du Tanzimat (1815-1871).\\
`AALI-PASHA, Turkish statesman, born in Constantinople. He attached his name to the policy of reforms of the Tanzimat (1815-1871).'
\end{quote}
has an unambiguous link to \verb=Q439237= as the occupation of the described person and the dates of birth and death match those of the wikidata item.

Some other definitions, like country names, are sometimes more difficult because the entities they describe can change in nature and form, while keeping the same name. As far as possible, we chose the wikidata item which is the closest to the Larousse entry in its historical context. For example, for
\begin{quote}
GRANDE-BRETAGNE et IRLANDE (Royaume-Uni de)...\\
`GREAT BRITAIN and IRELAND (United Kingdom of)...'
\end{quote}
we chose the \verb=Q174193= identifier, which describes a state that existed from 1801 to 1927. This is also the case for the municipalities, where we favored the historic municipalities over new ones resulting, for instance, from a merger.

Works of art are sometimes quite tricky, especially from Greek or Roman Antiquity, Middle Ages, or the Renaissance. The main reasons are changes of name or attribution. To solve certain cases, we had to resort to the \textit{Nouveau Larousse illustré}, a larger and earlier dictionary published between 1897 and 1904. This \textit{Nouveau Larousse illustré} contains more precise descriptions and sometimes a reproduction of the work in a figure, while keeping, for most entries, names identical to those in the \textit{Petit Larousse illustré}. This enabled us to identify many wikidata items.

Some entries contain lists of people or works of art, as for example:
\begin{quote}
ABAD Ier [bad'], premier roi maure de Séville, et chef de la dynastie des Abadites~; il régna de 1023 à 1042. — Son fils Abad II régna de 1042 à 1069, et son petit-fils, Abad III, de 1069 à 1095.\\
`ABBAD I [bad'], first Moorish king of Seville, and head of the Abbadid dynasty; he reigned from 1023 to 1042. — His son ABBAD II reigned from 1042 to 1069, and his grandson, ABBAD III, from 1069 to 1095.'
\end{quote}
Most of the time, such lists contain a dash character `--' to separate the entities. Using this character, we extracted all the lists and we connected all the names mentioned in them. In our annotation file, we also used a list to store the corresponding wikidata numbers. For the previous example, we linked the three entities to these identifiers: 
\begin{verbatim}
["Q305795", "Q30556", "Q299578"]. 
\end{verbatim}
At the end, the complete annotation of the entries resulted in 22,357 links, where 18,905 entries have one link and 1340 have two or more. Nonetheless, when an entry contains more than one name, the description text is usually much shorter from the second one. Consequently, in the subsequent experiments, we only considered the first identifier in the list.

\section{Interannotator Agreement}
We carried out an evaluation of the interannotator agreement on 200 randomly selected entries. See the list in Table \ref{ref:tab1} in the Appendix. The two annotators disagreed on 15 entries; the rest was identical. Out of these 15, two wikidata identifiers had been merged by wikidata editors between the first and second annotation runs, resulting in a difference in 13 entries. This corresponds to $\kappa$ coefficient \cite{Cohen1960} of 0.92, denoting a high agreement, and even 0.93, when removing the two synonyms.

We examined the 13 differences, for which we retained one of the two candidates. In this paper, we describe three of them, \textit{Caucase} `Caucasus', \textit{Exode} `Exodus', and \textit{Mariannes} `Marianas', to show the possible sources of confusion and the rationale to reconcile the disagreements.

\paragraph{Caucase.} The \textit{Caucase} entry has the following definition:
\begin{quote}
CAUCASE, chaîne de montagnes entre la mer Noire et la Caspienne...\\
`CAUCASUS, mountain range between the Black and Caspian Seas...'
\end{quote}
where the annotators proposed \verb=Q5477= and \verb=Q18869=. The first identifier refers to the Wikipedia article \textit{Caucasus Mountains}, while the second to \textit{Caucasus} as a region. Given the \textit{Larousse} description, we selected the first identifier.

\paragraph{Exode.} The \textit{Exode} entry is
\begin{quote}
Exode (l'), nom donné au deuxième livre du Pentateuque...\\
`Exodus (The), name given to the second book of the Pentateuch...'
\end{quote}
The annotators used the \verb=Q9190= and \verb=Q1290338= identifiers. The first one corresponds to the Book of Exodus while the second one links to an analysis the narrative of the Exodus as a putative historical event, told notably in the Book of Exodus, and described as a \textit{founding myth} in wikipedia\footnote{\url{https://en.wikipedia.org/wiki/The_Exodus}, first sentence, accessed December 3, 2021}. The \textit{Larousse} dictionary refers explicitly to the Book and we preferred the first identifier.

\paragraph{Mariannes.} The \textit{Mariannes} entry is
\begin{quote}
MARIANNES (îles), ou îles des LARRONS, archipel allemand du Pacifique...\\
`MARIANA (islands), or LADRONES Islands, German archipelago in the Pacific...' 
\end{quote}
The annotators linked the entry to \verb=Q16644=, corresponding to the Northern Mariana Islands and \verb=Q153732=, corresponding to the Mariana Islands. The Marianas were part of the Spanish Empire until the United States took over Guam, i.e. the southern part of the Marianas, in 1898, and the German Empire bought the northern part from Spain in 1899. Both parts are now U.S. territories. Again, we followed the definition and we linked the entry to the Northern Mariana Islands. Nonetheless, this case was more tricky as the definition in the \textit{Petit Larousse illustré}, following that of the \textit{Nouveau Larousse illustré}, is possibly erroneous, or at least incomplete.

Our dataset would certainly benefit from a complete double annotation. However, we did not have the resources to undertake it for the whole dictionary. Extrapolating our differences to the total number of entries, we could estimate the number of disagreements to 1316. Splitting them equally between the annotators, this means that 658 annotations (3.25\%) would need a revision.

\section{Dataset Format}
We stored the annotation in a JSON file as a list of objects equivalent to Python dictionaries. Each object represents an entry and has two keys: the raw text of the entry and a list of wikidata identifiers, most often only one. For the  AALI-PACHA entry, this corresponds to:
\begin{verbatim}
{"texte": "AALI-PACHA, homme 
 d'Etat turc, né à Constantinople. 
 Il a attaché son nom à la 
 politique de réformes du Tanzimat 
 (1815-1871).", 
"qid": ["Q439237"]}
\end{verbatim}

%JSON files are easy to share and process across most programming languages.

\section{Using the Wikidata Identifiers: The Semantic Categories}
\label{sec:types}
Using the identifiers, we can extract abundant information from wikidata. In this section and the next ones, we describe a few properties we can attach to the dictionary entries.

The first property of a wikidata item is its semantic class: The \textit{instance of} property. Its resembles Aristotle's categories or those in Wordnet \cite{Miller1993}, but it is much more comprehensive. 
%The Larousse features a significant number of short biographies, which  corresponds to the \textit{human} category in wikidata  with the \verb=Q5= identifier.

We extract this category from wikidata's graph using the SPARQL query language that consists of triple patterns like this one:
\begin{verbatim}
wd:Q439237 wdt:P31 ?type .
\end{verbatim}
where \verb=wd:Q439237= is Aali-Pacha's identifier, \verb=wdt:P31= is a property meaning \textit{instance of}, and \verb=?type= is the type we want to extract.

Querying a wikidata SPARQL server returns the semantic type of Aali-Pacha in this \verb=?type= variable: The \verb=Q5= identifier denoting a human. An entity may have more than one type, or have no type, if it is not documented in wikidata.

Table~\ref{tab:types} shows the ten most frequent types with their description.  The two first types are human with nearly 38\% of the entries and French commune with nearly 15\%.

As wikidata is continually updated by its users, classes may be added or removed. We will have then to rerun the queries to obtain the new values.

\begin{table}[t]
\centering
\begin{tabular}{lrp{3.5cm}}
\hline
Q-Number&Frequency&Description\\
\hline
Q5 &7653& human\\
Q484170& 3022&commune of France\\
Q1549591& 849& big city\\
Q515& 789&city\\
Q7725634& 572&literary work\\
Q4022& 447&river\\
Q3305213& 308&painting\\
Q747074& 255&\textit{comune} of Italy\\
Q1637706& 231&city with a population of more than 1,000,000\\
Q23442& 221&island\\
\hline
\end{tabular}
\caption{Wikidata semantic types of the entities in the \textit{Petit Larousse illustré}}
\label{tab:types}
\end{table}

As an alternative to wikidata's SPARQL endpoint, we could download the wikidata entry for \verb=Q439237= in the JSON format, \url{https://www.wikidata.org/wiki/Special:EntityData/Q439237.json}, and parse its content, or set up a private SPARQL server with the wikidata entries in the RDF Turtle format. The URL is the same, but with a \verb=ttl= suffix.

\section{The Human Beings in the \textit{Petit Larousse Illustré}}
Human being is the most frequent category and the corresponding wikidata items often provide the dates and places of birth and death of the person, sometimes the occupations, spouses, etc. Note that less known entities are not always completely documented.

In this section, we will focus on the dates of birth and death extracted from the \textit{Larousse} as well as from wikidata. 

\subsection{Information Extraction from Text}
The dates of birth and death are often present in the Larousse's entries and follow a quite regular pattern exemplified in the Aali-Pacha entry from Sect. \ref{sec:method}: They are located at the end of the description and nearly always shown as two years between parentheses separated by a dash. Using a very coarse strategy, we collected these dates when present in the definition with a simple regular expression:
\begin{verbatim}
[\(\[](\p{N}+)-(\p{N}+)\.?[\)\]]
\end{verbatim}

Applying it to all the entries, we could extract 5385 pairs of years.

\subsection{Information Extraction from Wikidata}
Similarly to the semantic types in Sect. \ref{sec:types}, we extracted the dates with the SPARQL triples:
\begin{verbatim}
wd:Q439237 wdt:P569 ?db .
wd:Q439237 wdt:P570 ?dd .
\end{verbatim}
where \verb=wdt:P569= is the property for the date of birth and \verb=wdt:P570= for the date of death.

Querying a server returns the values:
\begin{verbatim}
'1815-03-05T00:00:00Z',
'1871-09-07T00:00:00Z'
\end{verbatim}
from which we extract the years: 1815 and 1871.

The Wikidata queries for the human beings returned 7430 pairs. 

\subsection{Date Concordances Between the two Sources}
The pairs of dates obtained from text extraction and queries enabled us to verify the correctness of the annotation as well as complement the dictionary with up-to-date information.

In our dataset, 5144 entries had date pairs both from text and from wikidata. Out of them, 3957 matched exactly, corresponding to an accuracy of 77\%.

We looked at the rest, 1187 entries, and for most of them, the difference did not exceed one or two years. The entry:
\begin{quote}
ACUNHA DE FIGUEROA (don Francisco), poète américain, né à Montevideo (Uruguay) [1790-1862].\\
`ACUNHA DE FIGUEROA, (don Francisco), American poet, born in Montevideo (Uruguay) [1790-1862].'
\end{quote}
gives the dates 1790 and 1862, while wikidata proposes 1791 and 1862.

Introducing a margin in the match, only 339 entries showed a difference of more than five years. These remaining differences came from larger uncertainties in the date of birth or a wrong assumption on the date format in the \textit{Larousse} text. For example, certain pairs correspond to the first and last regnal years of kings, queens, or rulers as in:
\begin{quote}
TIBÈRE, deuxième empereur romain, fils de Livie et fils adoptif d'Auguste ;... (14-37).\\
`TIBERIUS, second Roman emperor, son of Livia and adopted son of Augustus;... (14-37).'
\end{quote}

\subsection{Distribution Across the Years}
\begin{figure}[t]
\centering
\includegraphics[width=\columnwidth]{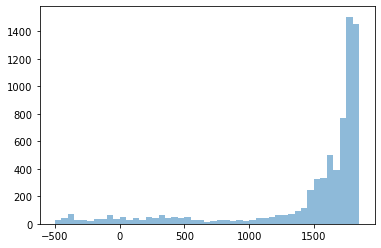} 
\caption{Distribution of birth dates in the \textit{Petit Larousse illustré} starting at year 500 BC. Bins of 50 years}
\label{fig:naissances}
\end{figure}

Excluding the dates before -500 and after 1851, and inspired by \newcite{escarpit1958}, we computed the distribution of the birth dates of the humans in the \textit{Larousse}, see Fig.~\ref{fig:naissances}. As source, we used the dates from wikidata as they are more complete than in the Larousse text, and possibly slightly more reliable. The two largest bins are between 1750 and 1800 and 1800 and 1850. These two most recent bins are twice as large as the one just before them between 1700 and 1750. 

An explanation of these figures lies in the growth of book production and printing industry in the 18th and 19th centuries, leaving easily available written traces of people's lives. It also clearly reflects the dictionary focus on contemporary people, mitigated by a modest, but noticeable number of humans from the classical antiquity. This emphasis on Classics, especially on Latin, is a feature of this dictionary, and beyond in high school education in France at the turn of the 20th century. For a discussion on this, see \newcite{chervel2008}.

\begin{figure}[t]
\centering
\includegraphics[width=\columnwidth]{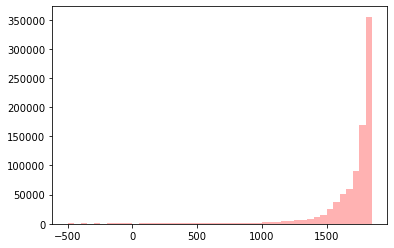} 
\caption{Distribution of birth dates in wikidata starting at year 500 BC. Bins of 50 years}
\label{fig:naissances_wd}
\end{figure}

As a comparison, wikidata contains a bit more than 9.5 million humans and we can contrast Fig.~\ref{fig:naissances} with Fig.~\ref{fig:naissances_wd} which shows the distribution of birth dates in wikidata. We observe a similar exponential growth, but between -500 and 500, and relatively to more recent years, the bars are hardly visible.

\begin{figure}[t]
\centering
\includegraphics[width=\columnwidth]{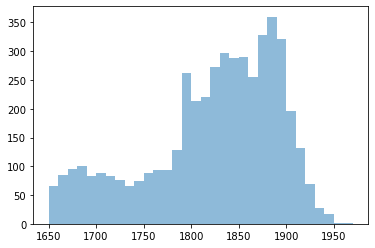} 
\caption{Distribution of death dates in the \textit{Petit Larousse illustré} starting at year 1650. Bins of 10 years}
\label{fig:morts}
\end{figure}

We can also examine the death dates in the \textit{Larousse} in Fig.~\ref{fig:morts}, this time within a smaller range and a distribution starting at year 1650. This again confirms the accent on contemporary or near-contemporary people. At the date of publication, 334 people were still living (as they died in 1905 or after). In Fig.~\ref{fig:morts}, we can see that year 1800 is indisputably a turning point, reflecting a specific attention of the \textit{Petit Larousse illustré} on the new era established by the French Revolution of 1789. For a discussion on such statistics, but limited to literature, see \newcite{darnton1971}.

\begin{figure}[t]
\centering
\includegraphics[width=\columnwidth]{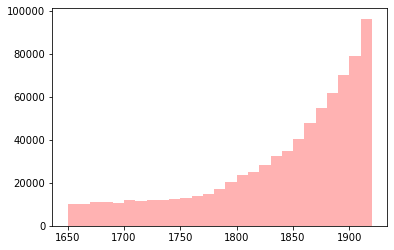} 
\caption{Distribution of death dates in wikidata starting at year 1650. Bins of 10 years}
\label{fig:morts_wd}
\end{figure}

As for the birth dates, we can compare this distribution with that of wikidata in Fig.~\ref{fig:morts_wd}, where the growth has a very regular pattern with no anomaly around year 1800.


\section{Geographical Entities}
\begin{figure*}[t]
\centering
\includegraphics[width=\textwidth]{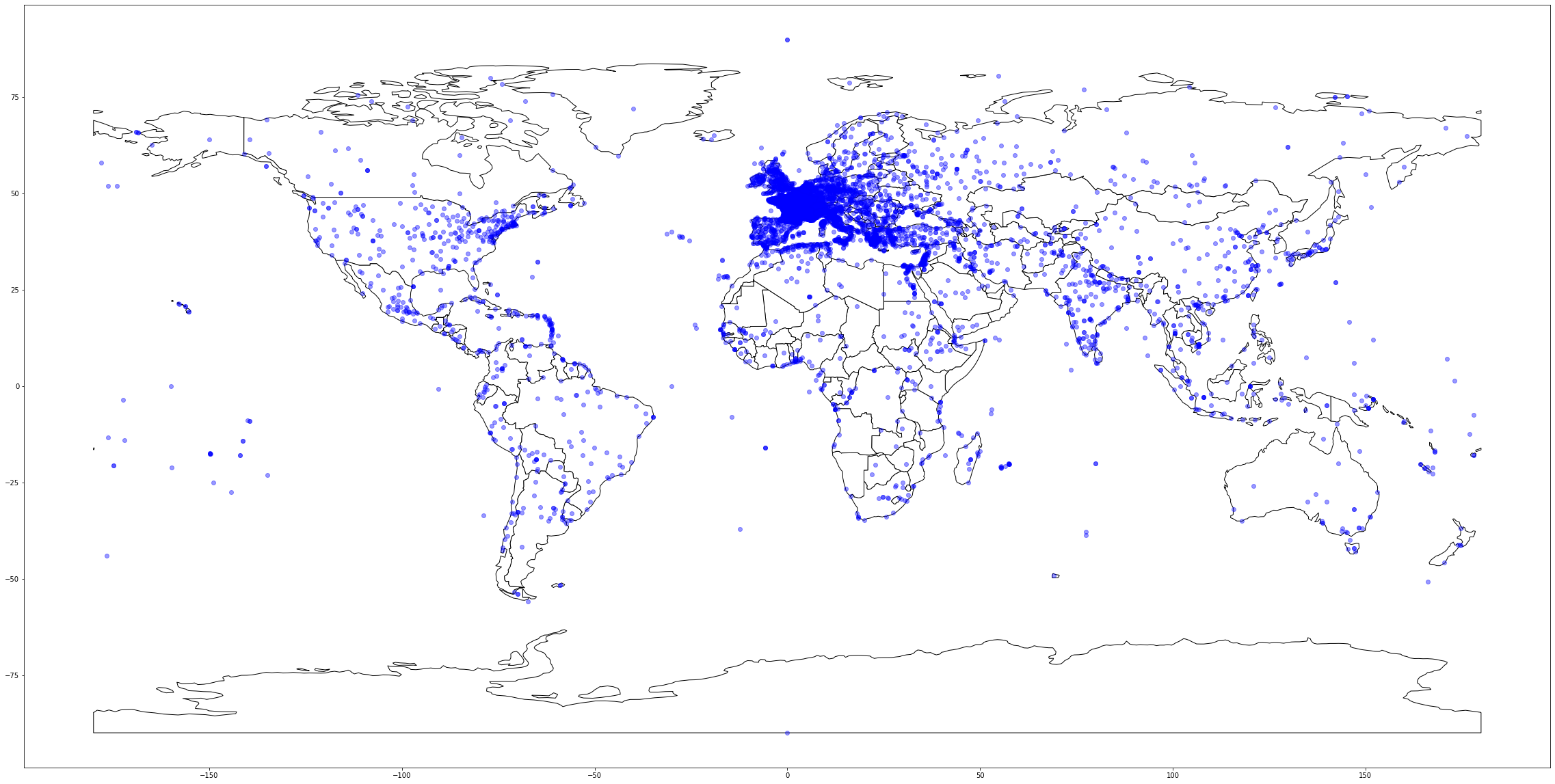} 
\caption{Geographic coordinates of the locations shown on a world map}
\label{fig:villes}
\end{figure*}

The geographical entities are another evidence of a world view conveyed by the Larousse. As for the types in Sect. \ref{sec:types}, we extracted the geographic coordinates from wikidata with the \verb=P625= property and the triple:
\begin{verbatim}
?e wdt:P625 ?geo .
\end{verbatim}

We applied the query to all the entities that have a location. This corresponds mainly to cities and towns, but also to countries, rivers, mountains, or any notable element of physical geography. Figure~\ref{fig:villes} shows the coordinates on a world map.

\section{Conclusion and Further Work}
%Finally, always due to lack of time, we could not have our annotations validated by other annotators. 
In this paper, we have described the annotation of all the entries of the history and geography part of the \textit{Petit Larousse illustré} from 1905 with their wikidata identifiers. We have also outlined how to use the identifiers to extract, complement, and process knowledge on the entities. This has enabled us to exhibit more precisely the scope of information and world view conveyed by this dictionary.

The analysis of historical encyclopedic knowledge is an open-ended research field and there are many issues we would like to explore. To name a few:
\begin{itemize}
    \item Compare the Larousse with other French encyclopedic works of the same time to analyze differences on geographic or human focus;
    \item Compare it with works from different periods to detect possible focus shifts;
    \item Compare the Larousse with encyclopedic works in other languages and from other countries and identify possible contrasts;
\end{itemize}

We have released the complete annotated corpus as well as code examples in the form of a Python notebook to analyze the links and extract information from wikidata.

We hope that this language resource will be useful to train entity linking applications or define new benchmarks and that it will facilitate new projects in the fields of lexicography, history, social sciences, and more generally digital humanities.

\section{Acknowledgments}
This work was partially supported by \textit{Vetenskaprådet}, the Swedish Research Council, registration number 2021-04533.
%This one to remove: 
%\citelanguageresource{speecon}

% \nocite{*}
\section*{Appendix}
Table \ref{ref:tab1} shows the list of entries used in the computation of the interannotator agreement. The indices refer to the position of the entries in the JSON file.

\begin{table*}[!ht]
\begin{center}
\begin{tabular}{llllllllll}
\hline
116&1779&3400&5003&8043&10663&12258&14564&16792&18454\\
123&1804&3487&5208&8093&10858&\textbf{12293}&14637&\textbf{16838}&18537\\
145&1827&3667&5478&8362&10900&12312&14652&16884&\underline{18675}\\
146&1984&3766&5633&8378&10981&12766&14796&17026&18724\\
294&2022&3778&5748&8626&11153&12780&14801&17063&18784\\
334&2130&3814&5825&8665&11196&12867&15194&\textbf{17071}&18977\\
665&2209&3828&6182&8746&11236&12929&15520&17256&\textbf{19029}\\
692&2255&3845&6384&8964&11241&12971&15533&17505&19136\\
741&\textbf{2271}&3848&6690&9096&11243&13221&15534&17731&19173\\
766&2333&3870&6875&9415&11324&13321&15577&17907&19237\\
795&2653&\textbf{3879}&\textbf{7170}&9489&\textbf{11359}&13444&15826&\underline{17954}&19246\\
823&2665&4209&\textbf{7370}&9667&11374&13537&15850&17968&19295\\
929&2850&4285&7403&9700&11489&13705&15897&18010&19493\\
990&3097&4411&7427&9824&\textbf{11604}&13875&16240&18113&19529\\
993&3099&\textbf{4457}&7494&10400&11683&13897&16402&18115&19567\\
1453&3140&4513&7561&10436&11775&13946&16440&18214&19645\\
1521&3169&4632&7572&10439&11776&14053&16554&18241&19700\\
1588&3194&4666&7869&10446&11797&14177&16584&18243&19924\\
1591&\textbf{3348}&4827&7877&10503&12194&14344&16609&18315&19936\\
1720&3359&\textbf{4979}&7899&10629&12207&14427&16734&18345&19993\\
 \hline
\end{tabular}
\end{center}
\caption{List of entries used in the evaluation of the interannotator agreement. The indices refer to the position in the json file. The numbers in bold correspond to entries, where the annotators disagreed. The underlined numbers correspond to identifiers that were merged between the two annotations}
\label{ref:tab1}
\end{table*}

\section{Bibliographical References}\label{reference}
%\label{main:ref}

\bibliographystyle{lrec2022-bib}
\bibliography{biblio}

%\section{Language Resource References}
\label{lr:ref}
\bibliographystylelanguageresource{lrec2022-bib}
\bibliographylanguageresource{languageresource}

\end{document}